\pdfoutput=1

\documentclass[11pt]{article}

\usepackage[]{emnlp2021}

\usepackage{times}
\usepackage{latexsym}
\usepackage{longtable}
\usepackage{graphicx}
\usepackage{makecell}
\usepackage{array}
\usepackage[T1]{fontenc}

\usepackage[utf8]{inputenc}

\usepackage{microtype}

\begin{document}
\title{Corporate Bankruptcy Prediction with Domain-Adapted BERT}


\author{Alex Gunwoo Kim$^*$\\
Seoul National University \\
Graduate School of Business \\
\texttt{kimgunwoo95@snu.ac.kr} \\\And
Sangwon Yoon$^*$\\
Artificial Society Inc. \\
\texttt{swyoon@artificial.sc}\\}

\maketitle
\def\thefootnote{*}\footnotetext{Equal contribution.}\def\thefootnote{\arabic{footnote}}
\begin{abstract} This study performs BERT-based analysis, which is a representative contextualized language model, on corporate disclosure data to predict impending bankruptcies. Prior literature on bankruptcy prediction mainly focuses on developing more sophisticated prediction methodologies with financial variables. However, in our study, we focus on improving the quality of input dataset. Specifically, we employ BERT model to perform sentiment analysis on MD\&A disclosures. We show that BERT outperforms dictionary-based predictions and Word2Vec-based predictions under time-discrete logistic hazard model, k-nearest neighbor (kNN-5), and linear kernel support vector machine (SVM). Further, instead of pre-training the BERT model from scratch, we apply self-learning with confidence-based filtering to corporate disclosure data. We achieve the accuracy rate of 91.56\% and demonstrate that the domain adaptation procedure brings a significant improvement in prediction accuracy.
\end{abstract}

\section{Introduction}

Predicting imminent corporate bankruptcies has been of great importance both in academia and in industry. Early studies on bankruptcy prediction focuses on identifying financial variables that precede impending insolvencies.  \citet{altman68} finds out that z-score, a composite measure of several financial variables, predicts imminent insolvencies. Since then, numerous papers document additional financial variables that seem to predict bankruptcies  (\citealp{ding12}; \citealp{bharath2008}; \citealp{dwyer2004}). Among 39 distinct financial variables, \citet{tian15} choose seven key variables that effectively predict bankruptcies within 12 months by LASSO.\\
However, in contrast to the fact that the majority of corporate disclosures contain non-financial information, textual disclosures have received relatively less attention. Following \citet{li2008}’s call for research on textual corporate disclosures, there have been numerous attempts (\citealp{tetlock2008}; \citealp{li2010}; \citealp{mayew15}) to analyze the textual sentiments of corporate disclosures. They commonly find that textual non-financial information has orthogonal informational value to the existing financial information. However, the majority of the analyses are based on the dictionary-based approach suggested by \citet{loughran11}.\\
In our study, we perform a BERT-based analysis on corporate disclosure data. BERT (\citealp{devlin2018}) is the pre-trained language model based on the self-attention mechanism of Transformers \cite{vaswani2017}. BERT and its improved versions such as GLUE \cite{wang2018}, SQUAD \cite{rajpurkar2016}, and RACE \cite{lai2017}, have achieved state-of-the-art results in several NLP downstream tasks. In this research, we analyze the management, discussion, and analysis (MD\&A) section of corporate disclosures and extract its context-specific sentiment. We then predict bankruptcies that occur within 12 months from the issuance of annual reports using the sentiment variables produced by BERT-based model. The reasons why we choose MD\&A sections to be our target of BERT-based analysis are as follows.\\
First, managers are obliged to express their opinions regarding the future performance of firms in MD\&A sections. Therefore, MD\&A is a rich source of information to analyze managerial assessment regarding a firm’s ability to operate as a going concern. Second, negative future predictions are likely to be accompanied by other positive explanations (see Appendix A, \citealp{jung1988}). Therefore, even though humans can interpret implicit negative nuance in the written disclosures, the traditional dictionary-based approach likely leads to an erroneous conclusion. Lastly, MD\&A sections are required when preparing 10-K filings for all firms. Therefore, we mitigate the sample selection bias by confining our analysis to observations archived in SEC filings.\\
Our paper makes several contributions to the existing line of literature. To our best knowledge, this is by far the first study to predict corporate outcomes other than stock market returns with BERT-based sentiment analysis. Unsophisticated investors have difficulty in understanding corporate disclosures since the disclosures are complex in nature \citep{bartov2000}. Therefore, the dictionary-based approach displays a trivial limitation in analyzing disclosure texts. We expect that context-specific linguistic analysis will accurately examine the contextual sentiment of corporate disclosures. Specifically, by comparing the ability to predict impending bankruptcies, we show that BERT-based analysis outperforms analyses based on dictionary (key word lists) and word level embedding.\\ 
Second, there is no BERT model trained on corporate disclosures and the open-source BERT model which is trained on the closest domain is Fin-BERT \citep{araci2019}. Fin-BERT is trained using financial news data. However, since the corporate disclosures and the financial news texts are in a similar but different domain, Fin-BERT is not perfectly suitable for interpreting corporate disclosures. We need to ensure that the data distributions of the training domain and the test domain are the same to improve the performance of machine learning models. Violation of this requirement, which is known as domain shift \citep{shimodaira2000}, leads to underperformance of models \citep{glorot2011}. Language models that are trained in two stages of pre-training and fine-tuning such as BERT, satisfy this assumption only when they are pre-trained and fine-tuned with a subset of their domain. Domain shift harms BERT model performance substantially (\citealp{lee2020}; \citealp{beltagy2019}). The most trivial way to overcome this problem is to pre-train BERT language model from scratch. However, language model pre-training is highly time and resource-consuming and it is inefficient to pre-train language models for only specific tasks.\\
Another way to deal with domain shift in BERT application is to fine-tune the model with labeled data from the target domain. However, in reality, labeled data for fine-tuning is often not available. In such cases, unsupervised domain adaption is a good alternative (\citealp{kim2020a}; \citealp{kundu2020}). In this paper, we apply self-learning, one of the key methodologies of unsupervised domain adaption. We show that if the distance between the source and target domains is close enough, supervising a BERT-based classification model with self-generated pseudo-labels filtered with confidence level leads to a significant improvement in performance.
\section{Related Studies}
\subsection{Bankruptcy predictions}
In his seminal study, \citet{altman68} finds that financial variables disclosed in annual reports predict bankruptcies. \citet{shumway2001} shows that in addition to financial statement-related variables, stock market-related variables such as market capitalization and stock price are also associated with future bankruptcies. However, considering that financial variables convey imperfect corporate information \citep{tennyson1990}, prior literature extracts information from narrative disclosures. \citet{cecchini2010} employ a complex vector space model to predict bankruptcies with MD\&A disclosures. However, they remain silent on whether textual information has additional predictive ability to financial variables. \citet{mayew15} find that narrative disclosures indeed contain information which is orthogonal to the information provided by financial variables. They utilize words lists provided by \citet{loughran11} to analyze general tone of MD\&A disclosures.\\
Related to prediction methodology, \citet{wilson1994} use neural network with financial variables to predict bankruptcies. \citet{premachandra2011} introduce data envelopment analysis (DEA) and show that bankrupt firms exhibit relatively lower operating efficiency. \citet{shin2005} find that SVM is effective in predicting notable corporate events including bankruptcies and \citet{chen2013} develop an adaptive fuzzy k-nearest neighbor method for insolvency prediction. Overall, prior literature has been successful in developing machine learning models that predict bankruptcies with considerable accuracy. However, few research focuses on improving the quality of input variables. Specifically, less effort has been made to produce precise semantic tone analysis with narrative disclosures.
\subsection{Text classification}
The most traditional method of text classification is the dictionary-based approach. The Harvard Psychological Dictionary is the most commonly used source in open domain text classification. \citet{loughran11} propose a dictionary specialized in the finance domain. However, dictionary-based approach has a limitation that it is difficult to create a dictionary that covers all the keywords needed for text classification and that the frequency of certain keywords does not necessarily contain sufficient information to classify sentences. Therefore, methods based on word embedding are suggested as alternatives.\\
Word embedding assigns a vector which encodes the meaning of the word to each text. Text classification methods based on word embedding include frequency-based methods as Tf-Idf \citep{salton1988} and prediction-based embedding methods as Word2Vec \citep{mikolov2013}. Word2vec, in particular, places each word in a vector space which approximates its semantic space. This algebraic transformation allows the vector operations among words. Therefore, we may set a word vector as the initial value of neural network and further classify sentences by exploring their contextual information. \citet{kim2014} prove that CNN structure, combined with Word2Vec embedding, can be used to classify sentences.\\
Language models based on Recurrent Neural Network(RNN) \citep{liu2016} and its variations (\citealp{zhou2015}; \citealp{wang2018}) are also used on text classification. However, recently, Transformer-based language models as BERT \citep{devlin2018} and GPT-2\citep{radford2019} outperform RNN-based methods and have drawn attention with their performance in generic benchmarks. These models apply self-attention to generate contextualized embedding. Especially BERT, the origin of many SOTA (State–Of–The–Art) models, pre-trains contextual embedding model with masked LM tasks and sentence prediction tasks, and is then fine-tuned to be applied to downstream tasks.
\subsection{Domain adaptation}
In its early stage, domain adaptation takes a form of semi-supervised learnings. Semi-supervised domain adaptation is used when labeled data exists in the target domain but when its amount is not sufficient. For instance, \citet{saenko2010} and \citet{kulis2011} use metric learnings to solve domain shifts. In specific, they adopt methods to learn task-specific distance metrics with labeled data and assign labels to unlabeled data based on the learned distance.\\
However, in reality, we may not be able to find domains with labeled data. In such a case, unsupervised domain adaptation (UDA) can be an attractive alternative. The subspace-based methods consider both source and target domain a subspace of single domain space. On the other hand, a more popular approach in UDA is to consider source and target domain separate spaces and try to align the distributions of these. Some works first compare the mean of samples from each domain in Hilbert space and assign a weight to each sample of source domain (\citealp{gretton2012}) or select samples in the source domain to minimize the maximum mean discrepancy of the two domains (\citealp{gong2013}). But when source and target domains are significantly different, we may not expect these methods to perform well. To deal with this problem, other studies (\citealp{pan2010}; \citealp{baktashmotlagh2013}; \citealp{sun2016}) map data from both domains to a latent space to deal with this problem.\\
Recently, with the advent of deep learning, feature extraction from raw data becomes an important process in every task. And the models that learn domain invariant features have become the mainstream in UDA (\citealp{ganin2016}; \citealp{saito2018}; \citealp{long2017}). However, these methods require the data from the source domain to extract domain invariant features. Therefore, self-learning can be an alternative since the source domain data is not required in the setting. The most important consideration in self-learning is how to generate or filter accurate pseudo-labels. \citet{kim2020a} propose confidence-based filtering and similarity-based pseudo-labeling method in the image classification task. However, their methodology cannot be directly applied to NLP tasks since word embedding is more implicit and multidimensional than image features. Recently, \citet{yoon2021} prove that fine-tuning the original model with pseudo-labels that are filtered based on confidence level increases accuracy in the target domain in the token classification task. To our best knowledge, our research is the first to show that self-learning without using samples from the source domain significantly improves model performance in sentence classification tasks. 
\section{Methodoloy}
\subsection{Sentiment analysis\footnote{For dictionary-based approach, we primarily utilize the following: \url{https://github.com/rflugum/10K-MDA-Section}. For the remainder, refer to our anonymized github: \url{https://anonymous.4open.science/r/BankruptcyBert-CC19/}}}
\subsubsection{Dictionary-based approach}
\citet{loughran11} develop word lists specifically suited for 10-K filings. They provide word lists that contain negative words and positive words, respectively. Following their methodology to calculate the tone of textual disclosures, we count the numbers of positive and negative words in each MD\&A section and scale them by the number of total words in each section (\textit{DICTPOS} and \textit{DICTNEG}). Although the analysis provides value-relevant information, the measures are comparatively less accurate in that they do not consider context-specific tone of the texts. We calculate the tone variables with Python.

\subsubsection{Word2Vec}
Word2Vec is a prediction-based word embedding method which trains by predicting center words with context words (CBoW) or vice versa (Skip-Gram). After training, each word in a corpus corresponds one-on-one to a vector that contains its semantic information. \citet{kim2014} achieve a remarkable performance on text classification by employing a structure with 1-layer convolutional neural network (CNN) and a fully connected output layer to classify sentences. This model takes pre-trained Word2Vec embedding as its input and the width of the filter in CNN equals the dimension of the word embedding. In our research, we replicate the CNN–static model of \citet{kim2014} in which the Word2Vec model freezes during the training. We use Word2Vec weight trained on the 10-K corpus of 1996-2013 \citep{tsai2016}, and train the network with the financial sentiment analysis dataset provided by \citet{malo2014} which consists of 4,846 sentences. The model takes each sentence as input and assigns probability to each of three classes: positive, negative, and neutral. We sum the probabilities of all sentences in a document and normalize them to calculate the sentiment score of each document (\textit{W2VPOS} and \textit{W2VNEG}). We use nltk sentence tokenizer\footnote{\url{https://www.nltk.org}} to split each document to sentences, gensim package\footnote{\url{https://radimrehurek.com/gensim/}} to load Word2Vec embedding, and Pytorch to implement CNN-based classifier. We use Cross-Entropy loss function and Adam optimizer. We train model for 60 epochs, with batch size 50. We set sentence length to 50 words in both training phase and inference phase.

\subsubsection{BERT}
BERT is a pre-trained language model with bidirectional transformers, which can be applied to downstream tasks after supervised fine-tuning with relatively low resources. We utilize the model structure based on the original BERT model \citep{devlin2018} and the fine-tuned weight of Fin-BERT \citep{araci2019} trained for financial sentiment analysis. Fin-BERT is pre-trained on the subset of Reuters TRC2 dataset which includes financial press articles and fine-tuned on the financial sentiment analysis dataset provided by \citet{malo2014}, which is identical to the dataset that we use to train the network of Word2Vec model. Similarly, the model takes each sentence as its input and assigns probability to each of three classes: positive, negative, and neutral. We sum the probabilities of all sentences in a document and normalize them to calculate the sentiment score of each document (\textit{BERTPOS} and \textit{BERTNEG}). Similarly, we use nltk sentence tokenizer  and Huggingface Transformers package\footnote{\url{https://huggingface.co}} with Pytorch to implement BERT. We set max sentence length to 512, which is the max length limit of BERT.

\subsubsection{Unsupervised domain adaptation}
\begin{figure}[t]
\centering
\includegraphics[width = 0.9\linewidth ]{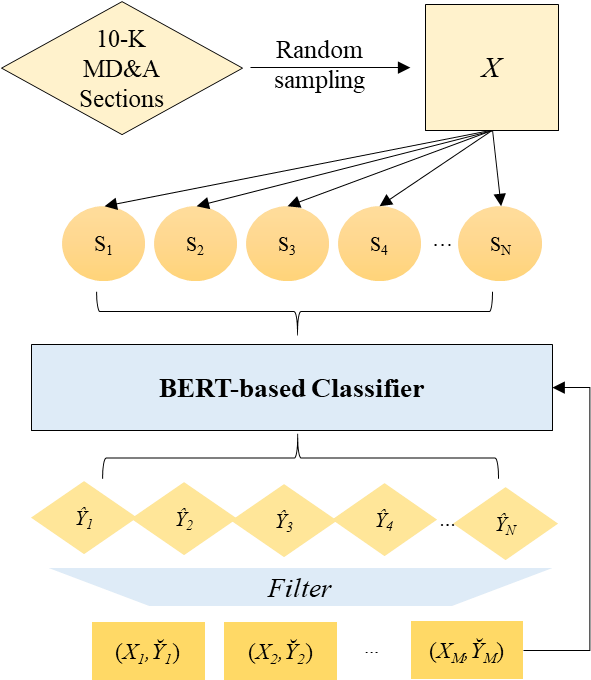}
\caption{This figure portrays the pipeline of our domain adaptation method. First, we randomly sample 1,200 documents from the corporate fillings from 1995 to 2020. We label this set of narrative disclosures $X$. Then we generate pseudo-labels $\widehat{Y}$ by applying a BERT-based classifier to every sentence in $X$ (denoted as $s_i$). To prevent noisy pseudo-labels from harming the model performance, we filter out only the ‘reliable’ samples ($X$, $\breve{Y}$) by their normalized self-entropy. Then we supervise the BERT-based classifier with reliable samples. }
\label{dapt}
\end{figure}

We apply self-learning, one of the unsupervised domain-adaptation methods, to our BERT-based model. The self-learning procedure follows the three-step approach. First, we generate pseudo-labels with sentences from MD\&A sections. Then we filter out “reliable” samples based on the self-confidence of the 
sentences. Since it is well known that erroneous labels may deteriorate the performance of the models, we filter the samples with high self-confidence. Specifically, we proxy for self-confidence with self-entropy (\citealp{zou2018}; \citealp{saporta2020}), Lastly, we perform supervised learning using the newly-obtained pseudo labels. Refer to Figure~\ref{dapt} for visual representation of the algorithm. We use the following equation to calculate self-entropy:
\[H(s_i) = - \frac{1}{\log M } \sum_{n=0}^{2} l_{n}(s_i)\log l_{n}(s_i)\]
,where $s_i$ denotes each sentence and $l_{n}(s_i)$ denotes the probability that $s_i$ belongs to class $n$ ($n=0,1,2$). Here, we calculate $l_n(s_i)$ with the BERT-based classification model that we use in section 3.1.3. Then we normalize the self-entropy by scaling the value with $\log M$, which is the natural logarithm of the number of labels (here, $M=3$). We define three classes 0, 1, and 2. Each class represents positive, negative, or neutral, respectively. 

We obtain 59,389 cleansed MD\&A filings merged with financial variables. Then we generate pseudo-labels by randomly picking 1,200 documents and doing inference on all sentences in the documents. In specific, we collect and analyze 589,858 distinct sentences. Next, we filter the results with the threshold of self-entropy 0.2 and discard the observations with self-entropy over 0.2. We obtain 38,703 reliable sentences through this process (6.56\% of the sentence domain). We train the model for 2 epochs with batch size 32 and set the learning rate to be $5e^{-5}$. We use Cross-Entropy loss and Adam optimizer. After training the model, the inference follows the same procedures as the previous two models and we generate sentiment variables \textit{DAPTPOS} and \textit{DAPTNEG}.
 
\subsection{Classification model}
We implement three basic machine-learning based classifiers (hazard time-discrete logistic model, SVM, and kNN) to evaluate the additional informativeness of textual data. Since models as DNN or RNN achieve state-of-the-art prediction accuracy only with financial variables, it is difficult to show the effect of adding textual variables. Therefore,  we compare the relative performance of the baseline classifiers to highlight the incremental prediction accuracy from adding BERT-based sentiment variables. 
\subsubsection{Hazard model}
We use proportional hazards model \citep{fine1999} to calculate prediction accuracy. \citet{shumway2001} finds that maximum log-likelihood estimation of discrete-time logistic regression yields consistent estimates. In specific, we estimate the following discrete-time logistic regression with maximum-likelihood estimation: \[\log h_i(t) = \beta X_i(t)\]
$h_i(t)$ refers to the risk of bankruptcy for firm $i$ at time $t$. $X_i(t)$ refers to a vector of firm $i$ at time $t$ that consists of variables that are known to precede bankruptcies. In our study $X$ includes financial variables and MD\&A sentiments. Specifically, we first run the regression with only financial variables (\textit{FIN}). Then we add dictionary-based variables (\textit{FIN}, \textit{DICTPOS}, and \textit{DICTNEG}), Word2Vec-based variables (\textit{FIN}, \textit{W2VPOS}, and \textit{W2VNEG}), BERT-based variables (\textit{FIN}, \textit{BERTPOS}, and \textit{BERTNEG}), and domain-adapted BERT-based variables (\textit{FIN}, \textit{DAPTPOS}, and \textit{DAPTNEG}), respectively. Using the obtained coefficients, we calculate the fitted values of $\log h_i(t)$ ($\widehat{\log h_i(t)}$) and classify the observation to be bankrupt if $\widehat{\log h_i(t)}>0.5$ and non-bankrupt otherwise. Continuous variables are winsorized at 1\% level to minimize the effect of outliers on regression results.

\subsubsection{k-Nearest Neighbors and Support Vector Machine}
To further enhance the classification performance, we employ k-nearest neighbors (kNN) and Support Vector Machine (SVM) algorithms, following prior literature. \\
kNN is a simple non-parametric classification method. First, we calculate the Euclidean distance between any two pair of observations. That is for observation vectors $X_1$ and $X_2$, we compute
\[d(X_1, X_2) = \sqrt{X_1\cdot X_2}\]
, where $\cdot$ denotes the inner product of the two vectors. Specifically, in our research, vector $X_i$ includes variables that precede insolvencies. We start from five financial variables and sequentially include sentiment variables calculated using different sentiment analysis models. Then, the algorithm computes the distances between observation $X_i$ which belongs to the test set and all other observations that belong to the train set. Next, it chooses $k$ smallest distances from the observation $X_i$ and label the distances. In our research, we set $k =5$. The algorithm classifies $X_i$ to be bankrupt if the number of bankrupt labels is greater than the number of non-bankrupt labels.\\
Next, SVM aims at finding a hyperplane that divides the dataset into distinct categories with the largest margin. Let $X_i$ be a training data and two classes are labeled with $y_i$ as either +1 or -1. We solve the following minimization problem with respect to hyperplane $w$:
\[\min \frac{1}{2}\left \| w \right \|^2+C\sum_{i=1}^M \xi_i\]
,where $y_i(wX_i+b)\geq1-\xi_i$. Here, $\xi$ denotes a slack variable and $C$ is a regularization parameter. In our setting, we set $C=1e^{-5}$. Further, we choose linear kernel for SVM classification. Linear kernel reduces the cost of computation but yields comparatively less accurate results. In our sample, univariate analysis results in section 4.1 indicate that linear kernel is acceptable to classify the dataset. We choose the most basic models of kNN and SVM since the primary purpose of our research is to compare the relative performance of text sentiment classification models.
\subsection{Accuracy calculation}
We report two distinct accuracy measures to compare the performance of our models. A1 is the ratio of the number of observations that are classified as non-bankrupt (CNB) under each model to the number of total non-bankrupt (NB) observations (A1 = CNB/NB). On the other hand, A2 is the ratio of the number of observations that are classified as bankrupt (CB) under each model to the number of total bankrupt observations (B) (A2 = CB/B).\\
For hazard model, we also report adjusted R-square (\citealp{nagelkerke1991}; \citealp{cox2018}).
\[R^2 = 1 - \exp\left[ {\frac{2\left(\log l(Fit) - \log l(Null)\right)}{n}}\right]\]
$\log l(Fit)$ and $\log l(Null)$ refer to the maximum log likelihoods of the fitted model and null model containing only the intercept term, respectively. Then, the equation can be rewritten as
\[- \log{(1 - R^2)} = 2 \left[\frac{\log l(Fit) - \log l(Null)}{n}\right]\]
\begin{table}[t]
\resizebox{0.48\textwidth}{!}{%
\begin{tabular}{lccc}
\hline
\textit{}        & \multicolumn{1}{l}{\textit{BRUPT=1}} & \multicolumn{1}{l}{\textit{BRUPT=0}} & Difference (t-stat) \\
\textit{}        & (1)                                  & (2)                                  & (1) - (2)           \\ \hline
\textit{WC}      & -0.5558                              & 0.2091                               & -0.7649***          \\
\textit{}        &                                      &                                      & (-3.91)             \\
\textit{RE}      & -0.3977                              & 0.0450                               & -0.4427***          \\
\textit{}        &                                      &                                      & (-3.47)             \\
\textit{SALE}    & 1.0161                               & 0.9607                               & 0.0554**            \\
\textit{}        &                                      &                                      & (2.15)              \\
\textit{MVE}     & 1.8010                               & 10.3129                              & -8.5119***          \\
\textit{}        &                                      &                                      & (-8.83)             \\
\textit{EBIT}    & -0.7172                              & 0.4907                               & -1.2079***          \\
\textit{}        &                                      &                                      & (-3.17)             \\
\textit{DICTPOS} & 0.0069                               & 0.0072                               & -0.0003***          \\
\textit{}        &                                      &                                      & (-3.36)             \\
\textit{DICTNEG} & 0.0182                               & 0.0136                               & 0.0046***           \\
\textit{}        &                                      &                                      & (17.79)             \\
\textit{W2VPOS}  & 0.3786                               & 0.4042                               & -0.0256***          \\
\textit{}        &                                      &                                      & (-5.53)             \\
\textit{W2VNEG}  & 0.0969                               & 0.0558                               & 0.0411***           \\
\textit{}        &                                      &                                      & (3.21)              \\
\textit{BERTPOS} & 0.1833                               & 0.2257                               & -0.0424***          \\
\textit{}        &                                      &                                      & (-12.04)            \\
\textit{BERTNEG} & 0.2433                               & 0.2140                               & 0.0293***           \\
\textit{}        &                                      &                                      & (9.41)              \\
\textit{DAPTPOS} & 0.1421                               & 0.2362                               & -0.0941***          \\
\textit{}        &                                      &                                      & (-13.21)            \\
\textit{DAPTNEG} & 0.3512                               & 0.1528                               & 0.1984***           \\
                 &                                      &                                      & (10.55)             \\
\textit{n}       & 520                                  & 58,867                               &                     \\ \hline
\end{tabular}%
}\caption{This table reports univariate analysis results. Column (1) reports the mean of the variables when \textit{BRUPT}=1. Column (2) reports the mean of the variables when \textit{BRUPT}=0. The last column reports the differences in mean values (Column (1) – Column (2)). We also report t-statistics that examine the statistical significance of the differences in the parenthesis. *, **, and *** indicate that the difference is statistically significant under 1\%, 5\%, and 10\% confidence level, respectively.}
\label{desc}
\end{table}
\begin{figure*}
\centering
\includegraphics[width=0.8\textwidth]{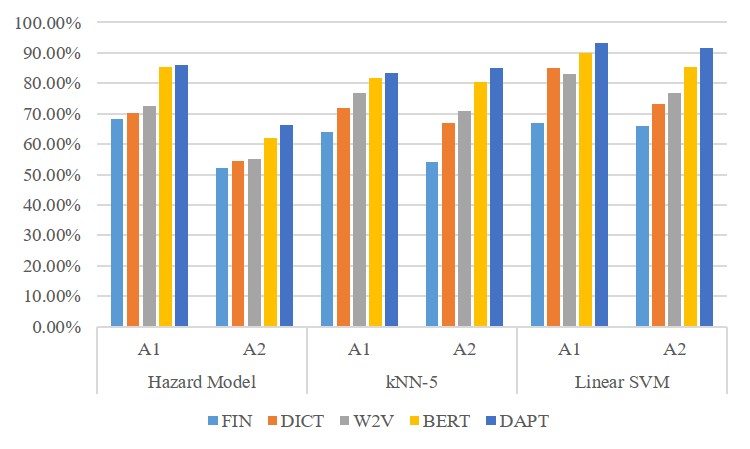}
\caption{This figure displays the relative accuracy prediction of the models. Refer to Section 3 for detailed definitions of \textit{FIN, DICT, W2V, BERT} and \textit{DAPT}. A1 measures the accuracy of predicting non-bankrupt firm-years and A2 measures the accuracy of predicting bankrupt firm-years. We display the prediction performance measured with hazard discrete logistic regression model, kNN-5, and linear SVM.}
\label{perf}
\end{figure*}
\section{Empirical Experiments}
\subsection{Data}
We first identify bankrupt firm-years from Compustat. The dataset provides us with the dates when firms file for bankruptcy and the dates when the bankruptcy procedure is complete. In our analysis, we use the dates when firms first file for bankruptcy as bankruptcy years. Then, following \citet{altman68} and \citet{mayew15}, we compute five key financial variables that are known to precede bankruptcies (\textit{WC, RE, EBITDA, MVE}, and \textit{SALE}). \textit{WC} refers to the ratio of working capital to total asset. \textit{RE} refers to retained earnings to total liability. \textit{EBITDA} refers to earnings before interest, tax, depreciation, and amortization scaled by total asset. \textit{MVE} is the market value of equity scaled by total liability. \textit{SALE} is the ratio of sales revenue to total asset. 
\begin{table*}[t]
\centering
\begin{tabular}{l|ccc|cc|cc}
\hline
\multicolumn{1}{c|}{\textbf{}} & \multicolumn{3}{c|}{Hazard Model} & \multicolumn{2}{c|}{kNN-5} & \multicolumn{2}{c}{Linear SVM} \\ \cline{2-8} 
Vars                           & R2        & A1        & A2        & A1           & A2          & A1             & A2             \\ \hline
\textit{FIN }                           & 16.23\%   & 68.20\%   & 52.23\%   & 63.88\%      & 53.99\%     & 66.83\%        & 65.82\%        \\
                               &           &           &           & (0.06)       & (0.05)      & (0.03)         & (0.03)         \\
\textit{FIN+DICT}                           & 21.33\%   & 70.25\%   & 54.55\%   & 72.01\%      & 66.98\%     & 85.11\%        & 73.28\%        \\
                               &           &           &           & (0.03)       & (0.03)      & (0.05)         & (0.06)         \\
\textit{FIN+W2V}                            & 23.51\%   & 72.39\%   & 55.23\%   & 76.75\%      & 70.93\%     & 83.20\%        & 76.73\%        \\
                               &           &           &           & (0.05)       & (0.04)      & (0.02)         & (0.04)         \\
\textit{FIN+BERT}                           & 24.83\%   & 85.44\%   & 62.00\%   & 81.73\%      & 80.32\%     & 89.98\%        & 85.20\%        \\
                               &           &           &           & (0.05)       & (0.05)      & (0.03)         & (0.06)         \\
\textit{FIN+DAPT}                           & \textbf{26.38\%}   & \textbf{86.12\% }  & \textbf{66.12\% }  & \textbf{83.22\% }     & \textbf{85.08\%}     & \textbf{93.26\% }       & \textbf{91.56\% }       \\
                               &           &           &           & (0.06)       & (0.04)      & (0.04)         & (0.02)         \\ \hline
\end{tabular}
\caption{This table reports the relative prediction accuracy of the models. We input the set of variables (\textit{FIN, DICT, W2V, BERT}, and \textit{DAPT}) in each of the three classification models. A1 equals (1 – Type I error rate) and A2 equals (1 – Type II error rate). We repeat the sampling experiment 100 times for each model and report the standard deviation of the accuracy rate in parenthesis.}
\label{comp}
\end{table*}
\\Next, we construct our main variables by extracting MD\&A sections from annual reports. Specifically, we inspect 10-K, 10-KSB, 10-K405, and 10KSB40 filings and search for MD\&A sections (Item 6 or Item 7). During the collection process, we exclude html notations, tables, and page numbers. This process ensures that we analyze only the textual components from MD\&A sections. Our sample period spans from 1995 to 2020 since SEC started to require firms to disclose electronic (machine-readable) filings from 1995. \textit{BRUPT} is an indicator variable that equals one for observations that face bankruptcy within 365 days from their issuance of annual report, and zero otherwise. We require all financial variables and MD\&A section texts for each observation and obtain 59,389 distinct observations. Among the sample, we identify 520 bankrupt firm-year observations (\textit{BRUPT}=1). We acquire financial data from Compustat database and filing texts from SEC archive.\\
To evaluate the performance of SVM and kNN models, we split the sample into three subsets: 60\% assigned to train set, 20\% to validation set, and 20\% to test set. However, since the data that we use is panel, randomly assigning 20\% of the sample to the test set may bias our results. That is, the model may learn from future information and use it to predict the same future. To mitigate this concern, we implement time-based split. That is, we choose 104 latest bankruptcy observations from 2018 to 2020 and randomly choose 104 non-bankruptcy observations from the same time period. To further ensure that our results are not driven by random sample selection, we repeat the selection procedure 100 times and report the average accuracy with its standard deviation.\\
In SVM, we test 10 different hyperparameters ranging from 0 to 1 and compare their relative performances. Next, for kNN model, we experiment five different hyperparameters ($k$). Since the model follows the majority rule, we examine odd parameters 3, 5, 7, 9, and 11. We then set the regularization parameter $C$ in SVM to be $1e^{-5}$ and the number of nearest neighbors $k$ to be 5. 
\subsection{Results}
To ensure that the selected variables move in accordance with \textit{BRUPT}, we report the univariate analysis results depending on the variable \textit{BRUPT} (See Table~\ref{desc}).\\
As evidenced by \citet{altman68} and other prior studies, we find higher \textit{WC, RE, MVE}, and \textit{EBIT}, and lower \textit{SALE} for non-bankrupt firms. Further, we demonstrate that \textit{DICTNEG, W2VNEG, BERTNEG,} and \textit{DAPTNEG} are higher for bankrupt firms and \textit{DICTPOS, W2VPOS, BERTPOS,} and \textit{DAPTPOS} are higher for non-bankrupt firms. This confirms that managers are likely to disclose negative-tone MD\&A sections before imminent bankruptcies. More importantly, untabulated tests including quadratic terms do not find any evidence that there exists non-linear relationship between \textit{BRUPT} and other independent variables. Taken together, univariate analysis results imply that we may choose linear kernel for SVM classification.\\
Table~\ref{comp} and Figure~\ref{perf} report the relative performance of the models. Consistent with the prior literature (\citealp{zhou2012}; \citealp{wu2007}), SVM generally performs the best among the three classifiers. Also, A1 is generally higher than A2 in all model specifications, implying that the models generally predict non-bankrupt firms more accurately than bankrupt firms.\\
Our main finding is that BERT-based analysis outperforms dictionary-based analysis and Word2Vec-based analysis. This indicates that context-specific sentiment analysis produces more accurate tone of the texts than non-context specific methods. Specifically, SVM with BERT-based sentiment variables display the bankruptcy prediction accuracy (A2) of 85.20\%. Further, we observe that R-square increases as we proceed from analyzing only financial variables (16.23\%) to including domain-adapted BERT-based sentiment variables (26.38\%). Comparing this result with prior literature that utilizes SVM to predict corporate bankruptcies, we obtain relatively high accuracy. For instance, \citet{zhou2012} obtain the accuracy rate of approximately 75\% by analyzing financial variables with DSSVM and GASVM models. Taken together, our results imply that textual information has predictive ability which is orthogonal to the existing set of financial variables and that adding high-quality textual information in the classifiers significantly improves the prediction accuracy.\\
Next, we also find that domain-adaptation further improves prediction accuracy. Domain-adapted BERT-based analysis yields the best accuracy rate (A2) of 91.56\% with linear SVM classifier among the models. These results strongly indicate that context-specific sentiment analysis of corporate disclosure texts provides more value-relevant information. 
\section{Conclusion}
In our study, we examine whether context-specific textual sentiment analysis (BERT) improves the accuracy of corporate bankruptcy prediction. We utilize five financial variables calculated from the stock market and annual reports that are known to precede impending insolvencies. Further, we collect and examine a large sample of MD\&A narrative disclosures from 1995 to 2020 to test whether textual sentiment is helpful in predicting financial distress. We find that textual sentiment has additional predictive ability to well-known financial variables. Most importantly, we show that BERT-based analysis outperforms dictionary-based analysis suggested by \citet{loughran11} and Word2Vec-based analysis combined with convolutional neural network. Further, we acknowledge the domain shifting problem of current BERT model. To assuage such a limitation, we perform domain-adaptation to the existing financial BERT model. This approach reduces computational costs compared to pre-training the BERT model with a new corpus and, at the same time, significantly improves the prediction accuracy.
\section{Acknowledgement}
The authors deeply appreciate helpful comments from Bok Baik and Yang Hoon Kim. Further, the authors appreciate the GPU support from Artificial Society.

\bibliography{anthology,custom}
\bibliographystyle{acl_natbib}

\appendix
\section{Sample MD\&A}
The following is an excerpt from an MD\&A section of 10-K report of Learning Tree International, disclosed on September 30, 1996. Learning Tree International filed a bankruptcy in 1997.\\
\newline
In response to the continued \textcolor{red}{\textbf{strength}} in enrollments, the Company has further accelerated its development of new course titles, \textcolor{red}{\textbf{expanded}} its future direct mailing plans to capture additional market share and has taken steps to expand the number of classrooms in its education centers. However, there can be \textcolor{blue}{\textbf{no assurance}} that the Company will be able to \textcolor{red}{\textbf{achieve}} an increase in market share after making such expenditures or will maintain its growth in revenues, \textcolor{red}{\textbf{profitability}} or market share in the future.
\newline
\\
Positive words are colored in \textcolor{red}{\textbf{RED}} and negative words are colored in \textcolor{blue}{\textbf{BLUE}}. Humans can interpret that this document conveys negative implication. However, there are 1 negative word and 4 positive words according to \citet{loughran11} word lists. In contrast, BERT-based sentiment vector of the paragraph equals (1.0365, 2.2161, 1.1704). Normalization yields \textit{BERTPOS} = 0.2343 and \textit{BERTNEG} = 0.5000. Therefore, BERT-based analysis outperforms the traditional dictionary-based approach.

\end{document}